\documentclass[10pt,twocolumn,letterpaper]{article}

\usepackage{iccv}
\usepackage{times}
\usepackage{epsfig}
\usepackage{graphicx}
\usepackage{amsmath}
\usepackage{amssymb}

\usepackage{graphicx}
\usepackage{amsmath}
\usepackage{amssymb}
\usepackage{booktabs}
\usepackage{xcolor}
\usepackage{multirow}
\usepackage{colortbl}

\usepackage[pagebackref=true,breaklinks=true,letterpaper=true,colorlinks,bookmarks=false]{hyperref}

\usepackage[capitalize]{cleveref}
\crefname{section}{Sec.}{Secs.}
\Crefname{section}{Section}{Sections}
\Crefname{table}{Table}{Tables}
\crefname{table}{Tab.}{Tabs.}

\iccvfinalcopy 

\def\iccvPaperID{****} 

\ificcvfinal\fi

\begin{document}

\title{Harnessing Low-Frequency Neural Fields for Few-Shot View Synthesis}

\author{
Liangchen Song$^{1}$\quad
Zhong Li$^{2}$\quad
Xuan Gong$^{1}$\quad
Lele Chen$^{2}$\quad
Zhang Chen$^{2}$\quad
Yi Xu$^{2}$\quad
Junsong Yuan$^{1}$\\
{ $^{1}$University at Buffalo \quad $^{2}$OPPO}
}
\maketitle

\begin{abstract}
  Neural Radiance Fields (NeRF) have led to breakthroughs in the novel view synthesis problem. Positional Encoding (P.E.) is a critical factor that brings the impressive performance of NeRF, where low-dimensional coordinates are mapped to high-dimensional space to better recover scene details. However, blindly increasing the frequency of P.E. leads to overfitting when the reconstruction problem is highly underconstrained, \eg, few-shot images for training. We harness low-frequency neural fields to regularize high-frequency neural fields from overfitting to better address the problem of few-shot view synthesis. We propose reconstructing with a low-frequency only field and then finishing details with a high-frequency equipped field. Unlike most existing solutions that regularize the output space (\ie, rendered images), our regularization is conducted in the input space (\ie, signal frequency). We further propose a simple-yet-effective strategy for tuning the frequency to avoid overfitting few-shot inputs: enforcing consistency among the frequency domain of rendered 2D images. Thanks to the input space regularizing scheme, our method readily applies to inputs beyond spatial locations, such as the time dimension in dynamic scenes. Comparisons with state-of-the-art on both synthetic and natural datasets validate the effectiveness of our proposed solution for few-shot view synthesis. Code is available at \href{https://github.com/lsongx/halo}{https://github.com/lsongx/halo}.
\end{abstract}

\section{Introduction}
Neural Radiance Field (NeRF) \cite{mildenhall2020nerf} and its extensions have shown promising results for novel view synthesis. In NeRF, coordinate-based multilayer perceptrons (MLPs), also referred to as neural fields, are adopted for continuously representing the geometry and appearance of the 3D scene. 
A critical factor that leads to the success of NeRF is Positional encoding (P.E.), which maps the low-dimensional coordinates to high-dimensional embeddings for representing the high-frequency details.

However, NeRF may generate unsatisfactory results when only a few posed images are available for training \cite{yu2021pixelnerf}. Under the few-shot view synthesis setting, the original NeRF overfits the input views and converges to a degenerate solution \cite{Jain_2021_ICCV}. When the radiance field is underconstrained, the high-dimensional P.E. in NeRF tends to fill the space with high-frequency geometry, resulting in poor generalization to unseen views. On the other hand, if we learn the neural field with low-frequency only P.E., fine details will be lost in the rendered views. 
Much significant progress has been made in adjusting the overall frequency of P.E. for a scene \cite{TancikSMFRSRBN20,AcornMartelLLCMW21,hertz2021sape,lindell2021bacon,Benbarka_2022_WACV}, however, in this paper, we argue that the potential behind low-frequency only neural fields can be further exploited for the few-shot setting.

\begin{figure}
    \centering
    \includegraphics[width=\columnwidth,trim={0 0 0 10pt},clip]{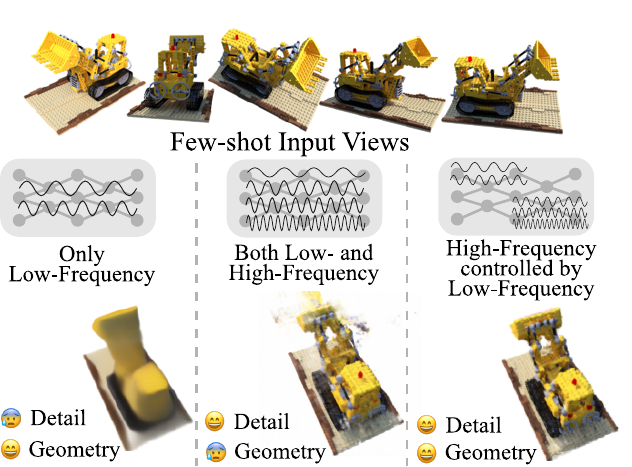}
    \caption{When solving the few-shot novel view synthesis problem with NeRF \cite{mildenhall2020nerf}, the frequency of positional encoding (P.E.) has a significant impact on its scene representation performance. Adopting a low-frequency P.E. for input coordinates leads to smooth geometry with fewer details, while adopting P.E. with both low- and high-frequency leads to overfitting and erroneous geometry. We propose to harness low-frequency neural fields for regularizing the high-frequency neural fields.}
    \label{fig:overview}
\end{figure}

\begin{figure}[t]
\centering
\includegraphics[width=0.8\columnwidth,trim={0 0 0 0pt},clip]{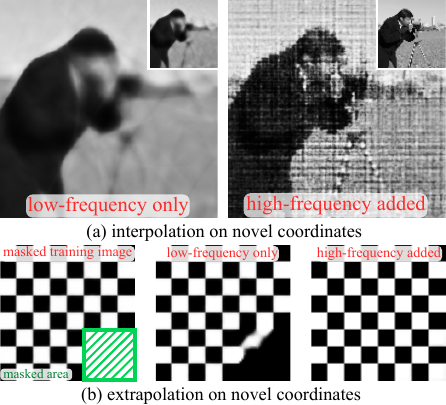}
\caption{A 2D toy example for demonstrating different interpolation and extrapolation results between a low-frequency only neural field and a high-frequency added neural field. The neural fields take as input the 2D coordinates of pixels and predict the intensities. (a) Given a $64^2$ image, the neural fields interpolate it to a $256^2$ image. The top right corner demonstrates the prediction on the training image after 10000 iterations. The low-frequency only neural field interpolates smooth values while the high-frequency equipped neural field generates obvious structured artifacts. (b) Given a checkerboard image with a corner masked, the neural fields extrapolate the masked area. The low-frequency only neural field cannot extrapolate the pattern to unobserved regions while the high-frequency added neural field succeeds.}
\label{fig:2dtoy}
\end{figure}

We are motivated by the difference between low-frequency only and high-frequency added neural fields on interpolation and extrapolation of novel input coordinates. Low-frequency only neural fields generate smooth interpolation but fail to extrapolate high-frequency periodic signals. In contrast, neural fields with sufficient frequency can adapt to periodic patterns but bring high-frequency interpolation artifacts (\cref{fig:2dtoy}).
An idea then arises that we can do interpolation with low-frequency only neural fields and extrapolation with high-frequency added neural fields. However, turning the idea into a practical algorithm is challenging since it involves classifying a prediction on novel coordinates as an interpolation or an extrapolation.

\begin{figure}
  \centering
  \includegraphics[width=0.8\columnwidth,trim={0 0 0 0},clip]{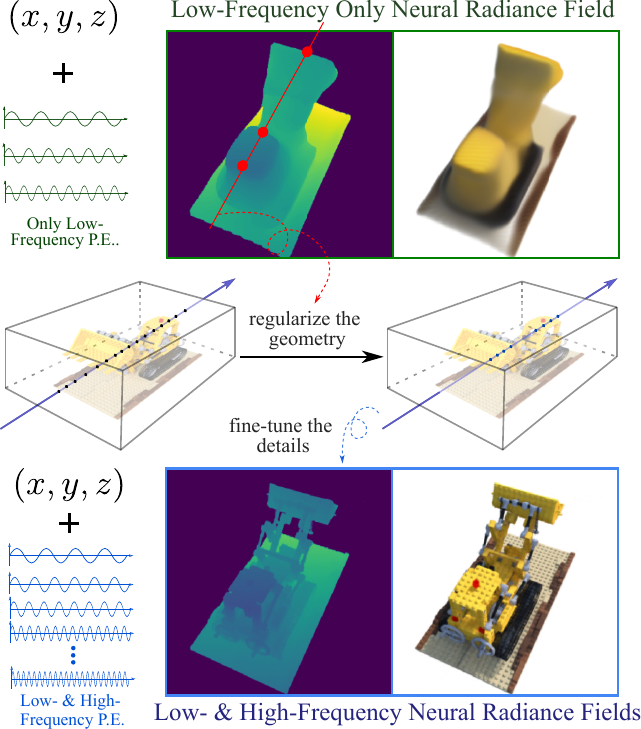}
  \caption{Controlling the geometry with a low-frequency neural field. The empty space of the two fields is enforced to be similar.}
  \label{fig:lf-illu}
\end{figure}

For the problem of few-shot synthesis, we treat the rough geometry prediction as the low-frequency interpolation only problem and the fine detail prediction as the high-frequency extrapolation needed problem. Our general idea is demonstrated in \cref{fig:overview}. Since querying the geometry information from point-based fields is expensive, we design a framework consisting of a low-frequency only ray-based field for modeling rough geometry and a high-frequency added point-based field for modeling fine details. The ray-based field is adapted from recent proposed light field networks (LFN) \cite{sitzmann2021lfns}, which has been proved far more efficient than point-based fields. During inference, the low-frequency only ray-based field will predict a depth value for each ray, instead of colors as in LFN. In this way, not only can the ray-based be trained easier since it only needs to fit a low-frequency smooth signal, but also the point-based radiance field can be efficiently regularized.

Furthermore, we propose to leverage the well-established frequency analysis of 2D image signals for determining the frequency of modeling rough geometry. Previous methods determine the frequency of neural fields by human or ground-truth-involved approaches \cite{YuCCXSY22}, which is not applicable in our few-shot view synthesis setting. We assume that slightly changing the viewpoint should not significantly impact the frequency domain of rendered 2D images if the scene is free of high-frequency floating artifacts. Based on the assumption, we set a threshold on the change in the frequency domain and then tune the frequency of P.E. from high to low until the rendered images satisfy the predefined threshold. The proposed criterion is simple but effective and does not require extra data.

For validating our proposed method, we first follow the 8 views setting and the 14 one side view setting used by DietNeRF \cite{Jain_2021_ICCV}. On this 360-degree rendering task, we train a low-frequency radiance field with the given eight views, and then it is used for supervising a ray-based field. 
Next, we consider a setting of 4 corner input views with challenging baselines on real data, which is practically appealing due to the complex geometry. 
In summary, our contributions are as follows:
\begin{itemize}
\item For few-shot novel view synthesis, we propose to harness the low-frequency inputs for regularizing the geometry, and then high-frequency inputs are added for details. Our regularization is conducted in the input space, enabling few-shot view synthesis on both static and dynamic scenes.
\item Benefiting from the different interpolation and extrapolation results between low-frequency only and high-frequency added neural fields, our framework can extrapolate periodic patterns in unobserved regions.
\item We propose a simple-yet-effective approach for tuning the frequency of P.E. under the few-shot view synthesis setting. The approach does not require extra data or labels, and the only hyperparameter (the threshold for the difference in rendered images' frequency domain) is well-generalizable across different scenes.
\end{itemize}

\section{Related Work}
\noindent\textbf{Neural fields and the frequency of inputs.}
Neural fields are also known as implicit neural representations or coordinate-based representations. 
Implicit neural representations are initially used for representing the geometry \cite{mescheder2019occupancy,saito2019pifu,park2019deepsdf,chen2019learning,chibane2020neural,jiang2020local}, but few works are concerned with high-frequency modeling details with a neural network. 
A recent milestone work in the field of novel view synthesis is NeRF \cite{mildenhall2020nerf}, in which a 5D field represents the scene. 
NeRF's impressive performance is largely due to the positional encoding module, which maps the input 5D coordinate to a higher dimensional space. The high-frequency mapping scheme is theoretically studied by FourierFeat \cite{TancikSMFRSRBN20} and SIREN \cite{sitzmann2020implicit}, in which the interpolation and extrapolation behavior with different input frequencies are explored as well.
For stabilizing training, the smooth prediction of a low-frequency prediction is leveraged in \cite{Park_2021_ICCV,lin2021barf}.
Coarse-to-fine frequency hierarchy is furhter adopted by Acorn \cite{AcornMartelLLCMW21}, BACON \cite{lindell2021bacon}, SAPE \cite{hertz2021sape} and \cite{yifan2022geometryconsistent}. 
The previous methods also demonstrate the benefits of low-frequency neural fields, but harnessing low-frequency neural fields has yet to be well exploited under the setting of few-shot view synthesis.
\\\noindent\textbf{NeRF for few-shot view synthesis.} 
A limitation of NeRF is that a certain number of input views are required for constructing a good radiance field. One reason is that the neural field is optimized on each scene separately and many excellent works are introducing extra supervisory signals for few-shot input views. 
PixelNeRF \cite{yu2021pixelnerf} proposes to condition the radiance field on the semantics of the input views. Learning a latent code for each scene has also been adopted by \cite{schwarz2020graf,wang2021ibrnet,henzler2021unsupervised}. GRAF \cite{schwarz2020graf} uses a discriminator on the rendered novel view patches. 
In \cite{tancik2021learned}, meta-learning is employed for learning a prior on the radiance field. Similarly, meta-learning has also been adopted for SDF \cite{SitzmannCTSW20} and light field \cite{sitzmann2021lfns,Feng_2021_ICCV}. 
Semantic priors learned from large-scale image or language datasets like CLIP are adopted as supervisory signals in DietNeRF \cite{Jain_2021_ICCV}. Depth or sparse point clouds reconstructed by classical methods are adopted in DSNeRF \cite{deng2021depth}, MVSNeRF \cite{Chen_2021_ICCV}, NerfingMVS \cite{wei2021nerfingmvs} and \cite{roessle2022depthpriorsnerf}. 
InfoNeRF \cite{kim2022infonerf} uses the entropy constraint of the density in each ray to minimize potential reconstruction inconsistency. RegNeRF \cite{niemeyer2022regnerf} proposes regularizing the geometry and appearance of patches rendered from unobserved viewpoints.
Different from most of the existing few-shot NeRF methods that regularize the model \textit{in the output space} (\ie, the rendered images), our regularization is conducted \textit{in the input space}. 
Concurrent work FreeNeRF \cite{yang2022freenerf} is also inspired by the phenomenon of overfitting with high-frequency inputs. They propose to associate the frequency with visible ratio, while in our work we first train with low-frequency only and then adding high-frequency.
The simplicity of our method enables readily apply the method to dynamic scenes that have not been studied in great detail. 
\\\noindent\textbf{Light field rendering with few-shot inputs.} 
Light field rendering is a widely studied and applied technology \cite{LevoyH96,GortlerGSC96}, and the angular-spatial resolution tradeoff is a long-standing problem. In \cite{LearningViewSynthesis}, the authors propose a learning-based approach for rendering novel views with four corner sub-aperture views from the light fields captured by the Lytro Illum camera. 
The concept of multi-plane images (MPIs) \cite{zhou2018stereo} is proposed for decomposing the observation views and learning view extrapolation. 
Predicting the depth of images for rendering novel views is also studied in \cite{choi2019extreme,tucker2020single,wiles2020synsin} with few inputs or only one input view.
While there is much progress on the few-shot inputs setting, there is much room for improving the rendering quality without using a different dataset or pre-trained network.

\begin{figure*}
    \centering
    \includegraphics[width=1.0\textwidth,trim={0 0 0 0},clip]{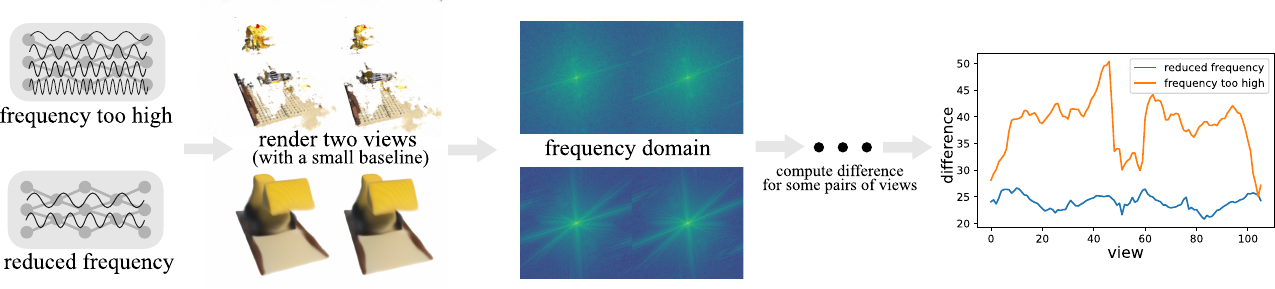}
    \caption{The proposed criteria for tuning the frequency of NeRF. A set of pairs of rendered views with a small baseline are considered. The difference of the pairs in the frequency domain is adopted as an indicator.}
    \label{fig:fourier-freq}
\end{figure*}

\section{Preliminaries}
In NeRF, a neural field $F_\Theta$ parameterized by an MLP that takes 3D point locations $\mathbf{p}=(x,y,z)$ and viewing direction $\mathbf{d}$ as input. 
The field is a mapping from each 3D location to its attributions $(\mathbf{c},\sigma)=F_\Theta(\mathbf{p},\mathbf{d})$, where $\mathbf{c}=(r,g,b)$ is the color and $\sigma$ is volume density. 
For rendering, each pixel color is acquired from points along the ray through the field $F_\Theta$ and differentiable volume rendering.
Each ray emitted from the camera center $\mathbf{o}$ with direction $\mathbf{d}$, points along the ray are $\mathbf{p}=\mathbf{r}(t)=\mathbf{o}+t\mathbf{d}$ and the expected color $C(\mathbf{r})$ with near and far bounds $t_n, t_f$ is
$C(\mathbf{r})=\int_{t_n}^{t_f} e^{-\int_{t_n}^t\sigma(\mathbf{r}(s)) ds}\sigma\big(\mathbf{r}(t)\big)\mathbf{c}\big(\mathbf{r}(t),\mathbf{d}\big)dt$.
A set of points are sampled along the ray for numerically estimating the integration. During training, the interval $[t_n, t_f]$ is partitioned into $K$ bins and for $i$th bin we uniformly sample a $t$ with
$t_i\sim U\left( t_n+\frac{i-1}{N}(t_f-t_n), t_n+\frac{i}{N}(t_f-t_n) \right)$.
Then the color of the ray is computed with
\begin{equation}\label{eq:int}
\hat{C}(\mathbf{r})=\sum_{k=1}^Ke^{-\sum_{k'=1}^{k-1}\sigma_{k'}\delta_{k'}}(1-e^{-\sigma_k\delta_k})\mathbf{c}_k,
\end{equation}
where $\sigma_k=\sigma(\mathbf{r}(t_k))$, $\mathbf{c}_k=\mathbf{c}\big(\mathbf{r}(t_k)\big)$ and $\delta_k=t_{k+1}-t_k$ is the distance between the two samples. 

The points $\mathbf{p}\in\mathbb{R}^3$ are low-dimensional and cannot represent the high-frequency details of the scene. In NeRF, the coordinate inputs to a higher dimension with positional encoding 
\begin{equation}\label{eq:pe}
\gamma(\mathbf{p}) = \left(\sin(\mathbf{p}),\cos(\mathbf{p}),\cdots,\sin(2^{L-1} \mathbf{p}),\cos(2^{L-1} \mathbf{p})\right),
\end{equation}
where $L$ is a hyperparameter. Positional encoding is critical for NeRF, but a high $L$ will lead to poor interpolation results, which has been theoretically studied by Tancik \etal \cite{TancikSMFRSRBN20}. Two fields, coarse and fine fields, are designed and trained jointly in NeRF. Low- and high-frequency P.E. are used for both two fields. The fine field takes samples near the visible contents predicted by the coarse field. 
Finally, the two fields are optimized with the reconstruction loss,
\begin{equation}\label{eq:rec}
L_{\mathrm{rec}} = \sum_{\mathbf{r}} \|\hat{C}(\mathbf{r})-C_{\mathrm{gt}}(\mathbf{r})\|^2_2,
\end{equation}
where $C_{\mathrm{gt}}$ is the ground truth RGB colors for ray $\mathbf{r}$. 

\section{Proposed Method}
The core idea of our method is to control the high-frequency components with a low-frequency only field.
We first present a criterion for determining the frequency of Lo-NeRF, then introduce a framework for efficiently regularizing high-frequency.
For simplicity, we denote the low-frequency only P.E. as \emph{Lo} and high-frequency equipped P.E. as \emph{Hi}. Lo-P.E. and Hi-P.E. equipped NeRF are denoted as Lo-NeRF and Hi-NeRF. 
\subsection{Frequency tuning}
Since we are concerned with the few-shot view synthesis setting, we design a criterion based on the rendered images' frequency domain and do not require extra data or labeling. An overview of the criteria is demonstrated in \cref{fig:fourier-freq}. Let $N$ paired images with a small baseline, denoted as $\{I^a_i,I^b_i\}$, are randomly rendered. Then the averaged difference on the pairs in the frequency domain is $\sigma=\frac{1}{N}\sum_i \|\mathcal{F}(I^a_i)-\mathcal{F}(I^b_i)\|$, where $\mathcal{F}()$ is Fourier transformation. Note that a mask is applied on the frequency domain to remove large values in the frequency domain for robustness (and detailed implementation can be found in supplementary). We reduce the frequency of NeRF until $\sigma$ reaches a predefined threshold (25 in our experiment). The frequency is then used for defining Lo-NeRF.
\subsection{Efficient low-frequency based regularizer}
We demonstrate in \cref{fig:overview} that Lo-NeRF produces a smooth geometry of the scene. A straightforward way of leveraging the smooth geometry from Lo-NeRF is separating the training into two stages: after training the Lo-NeRF, we enforce that the Hi-NeRF shares a similar rough geometry as the Lo-NeRF. The approach is illustrated in \cref{fig:lf-illu}. 
\paragraph{Low-frequency ray-based field.}
We propose to sample random rays in the space and then enforce the Hi-NeRF having consistent geometry for these random rays.
However, using a Lo-NeRF for regularizing can be time-consuming since querying the depth of a ray is rather expensive.
Thus, we adopt a ray-based field for regularizing the Hi-NeRF in an online manner: The computational cost can be reduced by directly predicting the depth value for each ray, as in DONeRF \cite{neff2021donerf}. Ray-based predictions may suffer from blurry edges in DONeRF. However, blurry predictions are acceptable in our case since only rough geometry is needed for regularization, and Hi-NeRF can fine-tune the details.

For training, the depth of a ray is first computed by sampling points and accumulates density outputs from the low-frequency point-based field. Then a ray-based field is supervised with the depth. All the ray origins $\mathbf{o}$ are the intersection between the ray at a predefined sphere and view direction $\mathbf{d}$ are unit vectors. So though the inputs of the ray-based field $(\mathbf{o},\mathbf{d})$ are six-dimensional, the degree of freedom is four since they are points from two surfaces.
Denote the ray-based field as $F_{\mathrm{ray}}$ and the point-based field (\ie, NeRF) as $F_{\mathrm{point}}$, then the loss for training $F_{\mathrm{ray}}$ is
\begin{equation}\label{eq:depth-sup}
L_{\mathrm{ray}} = \sum_{\mathbf{r}=(\mathbf{o},\mathbf{d})}\| F_{\mathrm{ray}}(\mathbf{o},\mathbf{d}) - D(\mathbf{r};F_{\mathrm{point}}) \|_2^2,
\end{equation}
where $D(\mathbf{r};F_{\mathrm{point}})$ is the depth computed with $F_{\mathrm{point}}$. 
\paragraph{Empty space loss.}
To regularize the empty space (\ie, where depth cannot be inferred from Lo-NeRF), we add an extra loss on the accumulated occupancy for a random ray $\mathbf{r}$, which is defined as $\mathrm{acc}(\mathbf{r})=\sum_{k=1}^Ke^{-\sum_{k'=1}^{k-1}\sigma_{k'}\delta_{k'}}(1-e^{-\sigma_k\delta_k})$. Denote the accumulated occupancy for the Lo-NeRF and Hi-NeRF as $\mathrm{acc}_{\mathrm{Lo}}$ and $\mathrm{acc}_{\mathrm{Hi}}$ respectively, then the empty space regularization loss is
\begin{equation}\label{eq:empty}
L_{\mathrm{empty}} = \sum_{\{\mathbf{r}:\mathrm{acc}_{\mathrm{Lo}}(\mathbf{r})<\tau\}} 
\mathrm{acc}_{\mathrm{Hi}}(\mathbf{r}),
\end{equation}
where $\tau$ is a threshold for determining if the space is empty, and we set $\tau=0.01$. Since accumulated occupancy values are all larger than 0, simply summing all the values up can effectively regularize the occupancy of all points on the ray to be 0.

\subsection{Overall framework}
As described above, our overall framework consists of 3 stages: Lo-NeRF training, ray-based field training, and Hi-NeRF training.
Lo-NeRF training is the same as training a typical NeRF except that the frequency of P.E. (\ie, $L$ in \cref{eq:pe}) is adjusted according to the frequency domain of rendered images. 
During the second stage, \ie, ray-based field training, we freeze the Lo-NeRF, and then random initialize a ray-based field. The ray-based field is then optimized with $L_{\mathrm{ray}}$ from \cref{eq:depth-sup}.
For the final stage, all previous networks are frozen, and we random initialize another NeRF with the default frequency in NeRF ($L=10$). The Hi-NeRF to be optimized will sample points according to the depth predicted by the ray-based field, following the same sampling strategy as in DONeRF. The training loss for the Hi-NeRF then becomes
\[
L=L_{\mathrm{rec}}+\lambda L_{\mathrm{empty}},
\]
where $\lambda$ is a balancing parameter. $L_{\mathrm{rec}}$ is computed by \cref{eq:rec} and $L_{\mathrm{empty}}$ is computed by \cref{eq:empty}.

\begin{table}[t]
\centering
\small
\begin{tabular}{lccc}
\toprule
\textbf{Method} & \textbf{PSNR}$\uparrow$ & \textbf{SSIM}$\uparrow$ & \textbf{LPIPS}$\downarrow$ \\\midrule
NeRF \cite{mildenhall2020nerf} & 14.934 & 0.687 & 0.318 \\
NV \cite{lombardi2019neural} & 17.859 & 0.741 & 0.245 \\
Simplified NeRF \cite{Jain_2021_ICCV} & 20.092 & 0.822 & 0.179 \\
DietNeRF \cite{Jain_2021_ICCV} & 23.147 & \underline{0.866} & \underline{0.109} \\
DietNeRF ft \cite{Jain_2021_ICCV} & \textbf{23.591} & \textbf{0.874} & \textbf{0.097} \\
HALO (\emph{Ours}) & \underline{23.269} & \underline{0.863} & 0.152 
\\ \cmidrule(lr){1-4}
HALO+DietNeRF & {23.687} & {0.879} & 0.090 \\
\textcolor{gray}{NeRF, 100 views} & \textcolor{gray}{31.153} & \textcolor{gray}{0.954} & \textcolor{gray}{0.046} \\
\bottomrule
\end{tabular}
\caption{Eight training views are randomly sampled for each scene on Realistic Synthetic scenes. Metrics averaged across the 8 scenes are reported.}
\label{tab:8view-compare}
\end{table}

\begin{figure}[t]
    \centering
    \includegraphics[width=1.0\columnwidth,trim={0 0 0 0},clip]{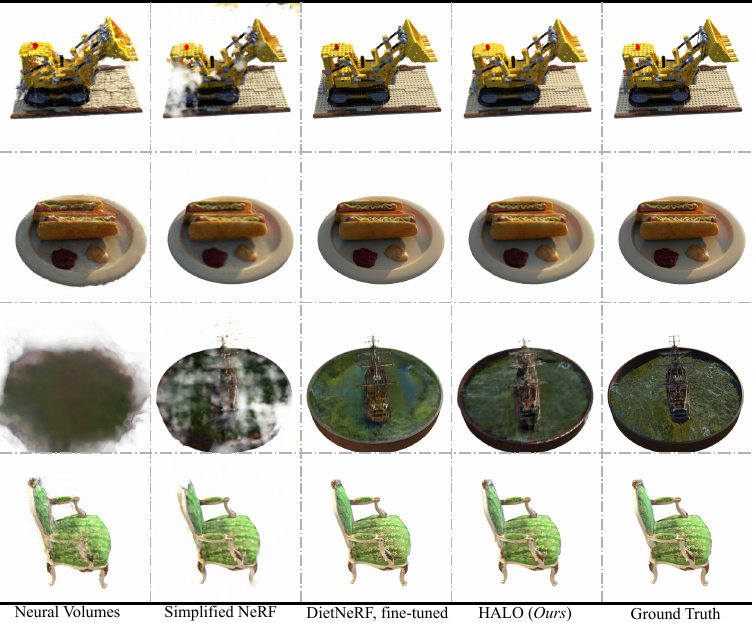}
    \caption{Comparison of novel views rendering results under the random 8 views setting.}
    \label{fig:8view-compare}
\end{figure}

\begin{table}[t]
\centering
\small
\begin{tabular}{lccc}
\toprule
\textbf{Method} & \textbf{PSNR}$\uparrow$ & \textbf{SSIM}$\uparrow$ & \textbf{LPIPS}$\downarrow$ \\\midrule
NeRF \cite{mildenhall2020nerf} & 19.662 & 0.799 & 0.202 \\
Simplified NeRF \cite{Jain_2021_ICCV} & 21.553 & 0.818 & 0.160 \\
DietNeRF \cite{Jain_2021_ICCV} & 20.753 & 0.810 & 0.157 \\
DietNeRF ft \cite{Jain_2021_ICCV} & \underline{22.211} & \underline{0.824} & \textbf{0.143} \\
HALO (\emph{Ours}) & \textbf{22.581} & \textbf{0.827} & \underline{0.150} \\ 
\cmidrule(lr){1-4}
HALO+DietNeRF & {22.862} & {0.833} & {0.140} \\ 
\textcolor{gray}{NeRF, 100 views} & \textcolor{gray}{31.618} & \textcolor{gray}{0.965} & \textcolor{gray}{0.033} \\
\bottomrule
\end{tabular}
\caption{A quantitative comparison under the one side views setting for the ``Lego'' scene. Note that DietNeRF requires a trained CLIP model \cite{radford2021learning} for supervision, while we do not.}
\label{tab:onside-compare}
\end{table}

\begin{figure*}[t]
    \centering
    \includegraphics[width=1.0\textwidth,trim={0 0 0 0},clip]{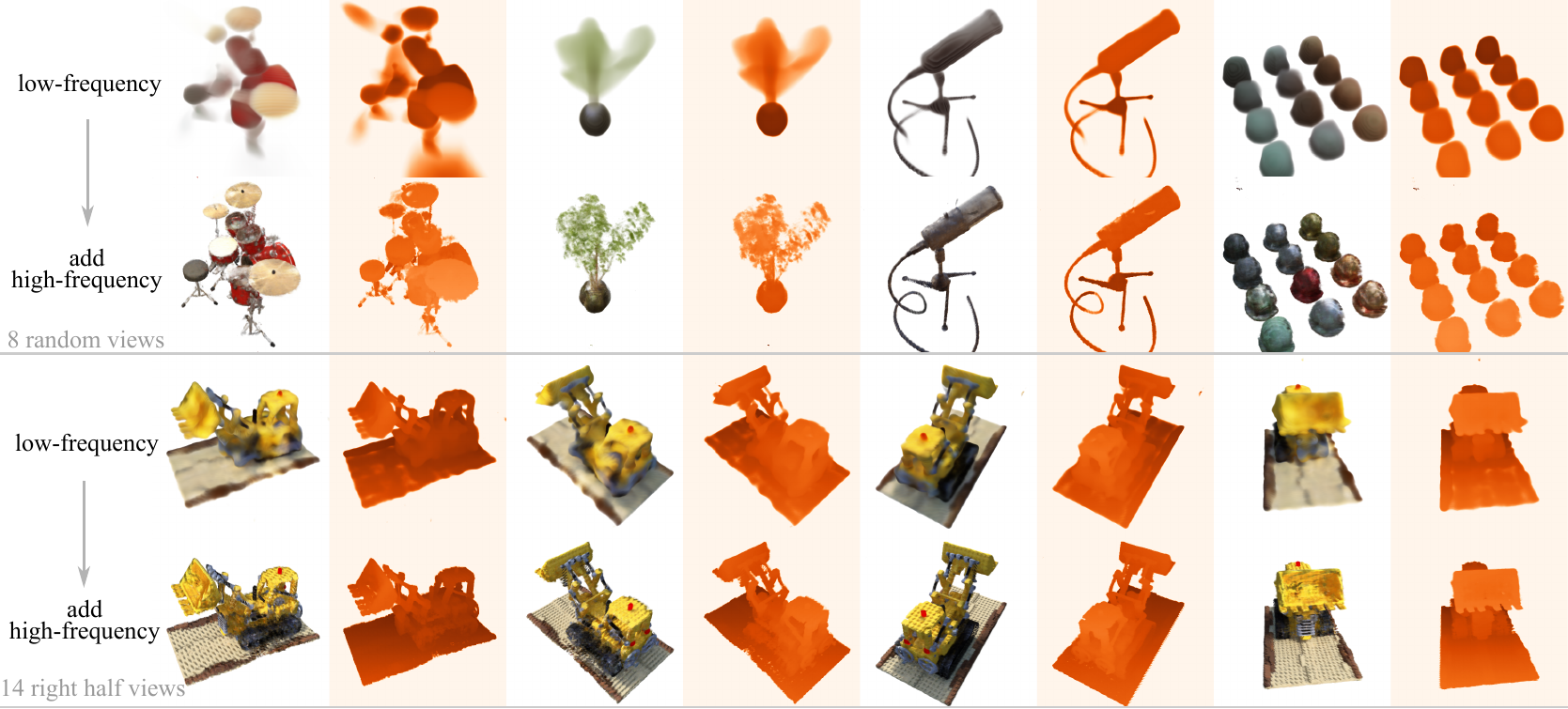}
    \caption{Novel views synthesized from the low-frequency neural field and the corresponding controlled high-frequency neural field.}
    \label{fig:lfhf-compare}
\end{figure*}

\begin{figure}[t]
    \centering
    \includegraphics[width=1.0\columnwidth,trim={0 0 0 5pt},clip]{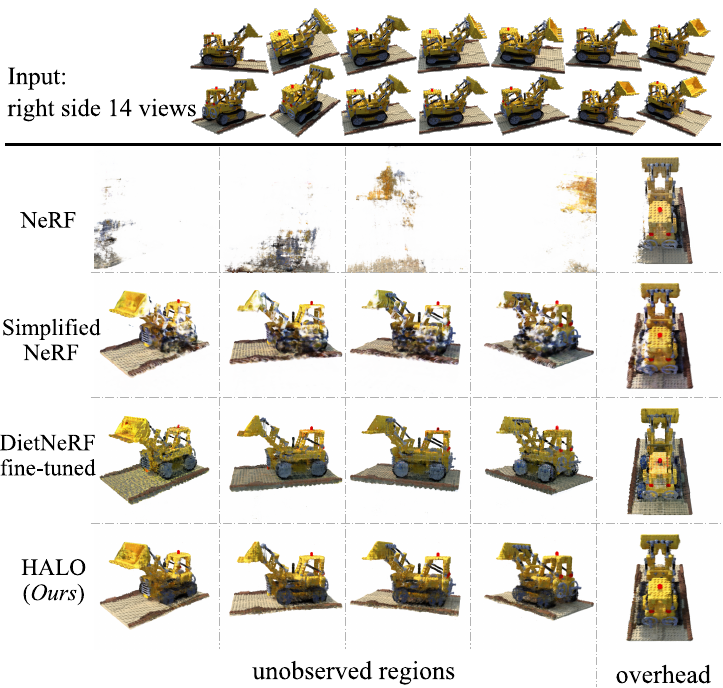}
    \caption{Novel views extrapolation with right side 14 views as inputs. The testing views are from the other side, which is not covered by the training views.}
    \label{fig:oneside-compare}
\end{figure}

\section{Experiments}
We validate our method in two aspects. First, we follow the few-shot settings in DietNeRF \cite{Jain_2021_ICCV}: 360$^\circ$ rendering from only 8 views and rendering on views not observed during training. Next, we present rendering results on a two-plane light field dataset with 4 corner views. Second, we test our method on dynamic scenes from D-NeRF \cite{pumarola2020d} with every 8 image from the original training set.
Our method is denoted as \textbf{HALO} (short for \textbf{HA}rnessing \textbf{LO}w-frequency).
Detailed implementation is described in the supplementary.

\subsection{Datasets and evaluation metrics.} For 360$^\circ$ rendering tasks, we use the Realistic Synthetic scenes \cite{mildenhall2020nerf}. 
There are 8 objects and images rendered from the objects are split into the train, validation, and test sets. We report results on the test set. 
Stanford Light Field Archive (StanfordLF) \cite{stanfordlf} is used to evaluate the two-plane light field data. 
There are 12 scenes, and each scene has $17\times17$ images. For each scene, we select the images with index (4,4) (4,12) (12,4) (12,12) as the four corner images. Rendering results on images indexed by (8,6) (8,10) (6,8) (10,8) are compared to ground truth for evaluation. 
The LLFF \cite{mildenhall2019llff,mildenhall2020nerf} dataset is also used for evaluating the performance of non-structured light field data. 
We manually select 4 views from 4 challenging scenes in LLFF as the training set, and an image inside the 4 views is used for evaluation. 
We use the DNeRF dataset \cite{pumarola2020d} and every eighth training image as the few-shot inputs for dynamic scenes. Different from static scenes, few-shot inputs on DNeRF indicate being sparse both spatial and temporal.

For evaluation, three metrics are used for evaluating the rendering results: Peak Signal-to-Noise Ratio (PSNR), Structural Similarity Index Measure (SSIM) \cite{wang2004image} and Learned Perceptual Image Patch Similarity (LPIPS) \cite{zhang2018unreasonable} with VGG backbone.

\begin{figure*}
    \centering
    \includegraphics[width=1.0\textwidth,trim={0 0 0 0},clip]{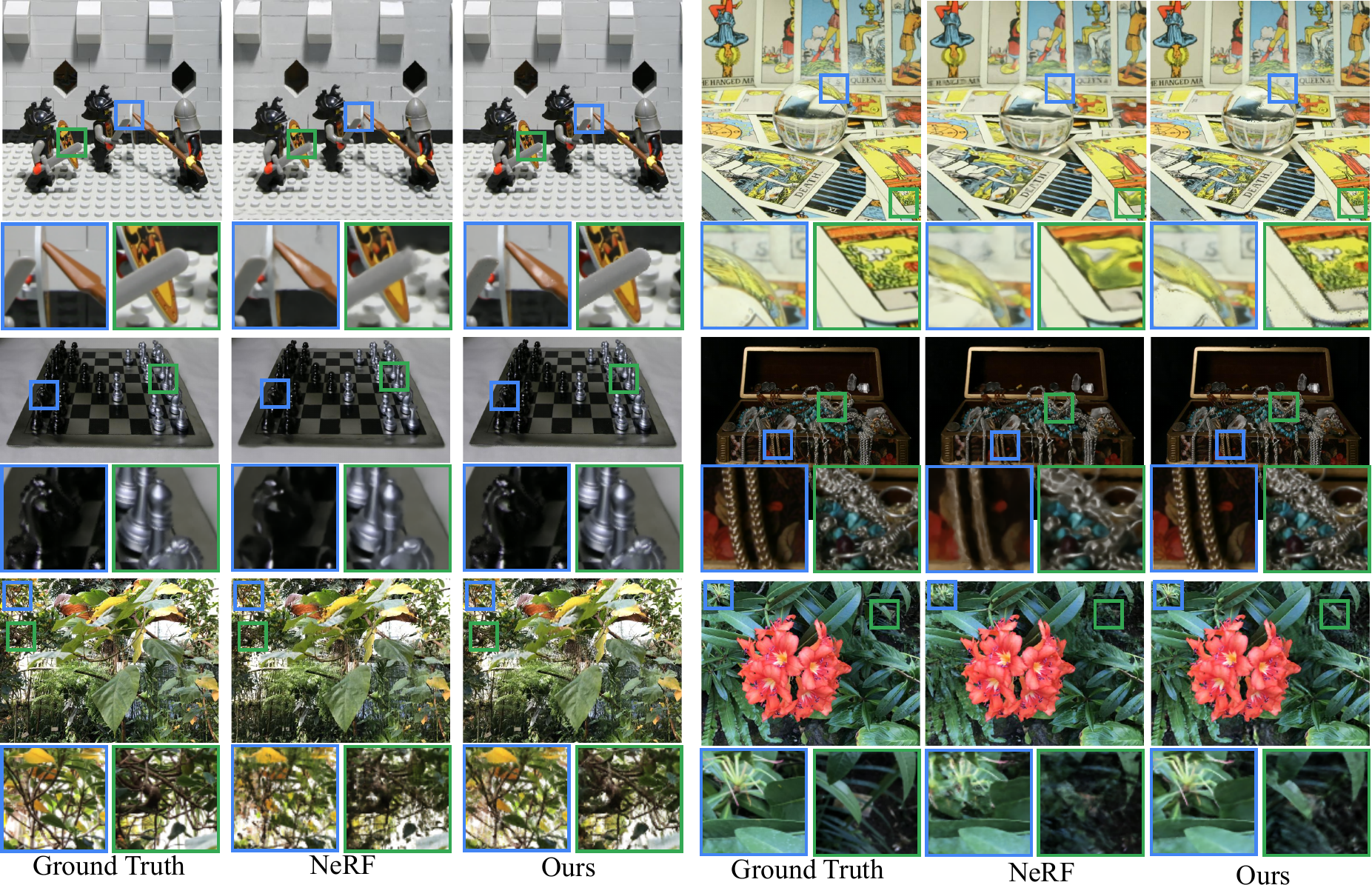}
    \caption{Qualitative results on StanfordLF (top \& middle rows) and LLFF (bottom row). The training views are four corner views, and the testing views are the center view. Despite being few-shot (4 views), the baseline is small, so NeRF can render decent images. The results demonstrate that harnessing low-frequency can improve details on few-shot inputs with small baselines.
    }
    \label{fig:image-compare}
\end{figure*}

\subsection{360$^\circ$ scenes}
We validate our method with the 360$^\circ$ rendering settings proposed by DietNeRF \cite{Jain_2021_ICCV} in this section. 
\paragraph{Random views as inputs}
To test the few-shot performance, 8 views are randomly selected by DietNeRF. We use the same random views for testing (detailed image names included in the supplementary). Overall performance on all scenes are reported in \cref{tab:8view-compare}. Our results achieve comparable results to DietNeRF without using extra semantic supervision signals. \cref{fig:8view-compare} visualizes the reconstruction results of the methods. An interesting point can be observed by comparing our method and DietNeRF on the ship object. DietNeRF-ft and our method generate decent results, but the brightness of our reconstruction is more consistent with the ground truth. This demonstrates that a potential drawback of introducing semantic supervision affects the lighting consistency, while our HALO can fill in the gap.
Further, our method and DietNeRF are orthogonal techniques and can potentially be combined. We report in the table a naive combination that first training with our method and then adding DietNeRF based fine tuning. The improvements demonstrate the potential of combining our method and other data-driven prior based methods. 
\paragraph{One side views as inputs}
In DietNeRF, the authors demonstrate the extrapolation ability enabled by introducing semantic supervision through reconstructing unobserved regions. We follow the setting of using 14 input views from the right side, and results are reported in \cref{tab:onside-compare} and \cref{fig:oneside-compare}. 
It is intriguing to observe that a neural field itself can well extrapolate the complex periodic signals since no extra supervisory signals are adopted in our method. 
Moreover, our method outperforms DietNeRF-ft in terms of PSNR and SSIM. By comparing the results in \cref{fig:oneside-compare}, we observe similar effects as the ship scene in \cref{fig:8view-compare}: the brightness of images reconstructed by DietNeRF is not consistent with the ground truth. 
This explains the reason behind the higher LPIPS score achieved by DietNeRF as the perceptual metric cares less about colors \cite{johnson2016perceptual}.

We also demonstrate the outputs from low-frequency and high-frequency neural fields in \cref{fig:lfhf-compare}. 
We can observe from the images that the extrapolation behavior between ours and DietNeRF is quite different: DietNeRF generates structurally consistent new contents, while ours extrapolate periodic contents of both structure and texture with high fidelity. Our method's artifacts are mainly on edges next to the wheels, and we attribute them to the complex structural signal with poor periodic patterns. 

\begin{table}
\centering
\small
\begin{tabular}{lccc}
\toprule
\textbf{Method} & \textbf{PSNR}$\uparrow$ & \textbf{SSIM}$\uparrow$ & \textbf{LPIPS}$\downarrow$ \\\midrule
\multicolumn{4}{l}{\emph{On Stanford Lightfield Archive} \cite{stanfordlf}}\\
NeRF \cite{mildenhall2020nerf} & 30.487 & 0.819 & 0.248 \\
HALO (\emph{Ours}) & \textbf{31.283}  & \textbf{0.897} & \textbf{0.234} \\ \cmidrule(lr){1-4}
\multicolumn{4}{l}{\emph{On LLFF} \cite{mildenhall2020nerf}}\\
NeRF \cite{mildenhall2020nerf} & 19.907 & \textbf{0.634} & 0.340 \\
HALO (\emph{Ours}) & \textbf{20.578} & {0.627} & \textbf{0.322} \\ 
\bottomrule
\end{tabular}
\caption{Quantitative comparison on forward-facing scenes.}
\label{tab:real-images}
\end{table}


\begin{figure*}[t]
\includegraphics[width=\textwidth]{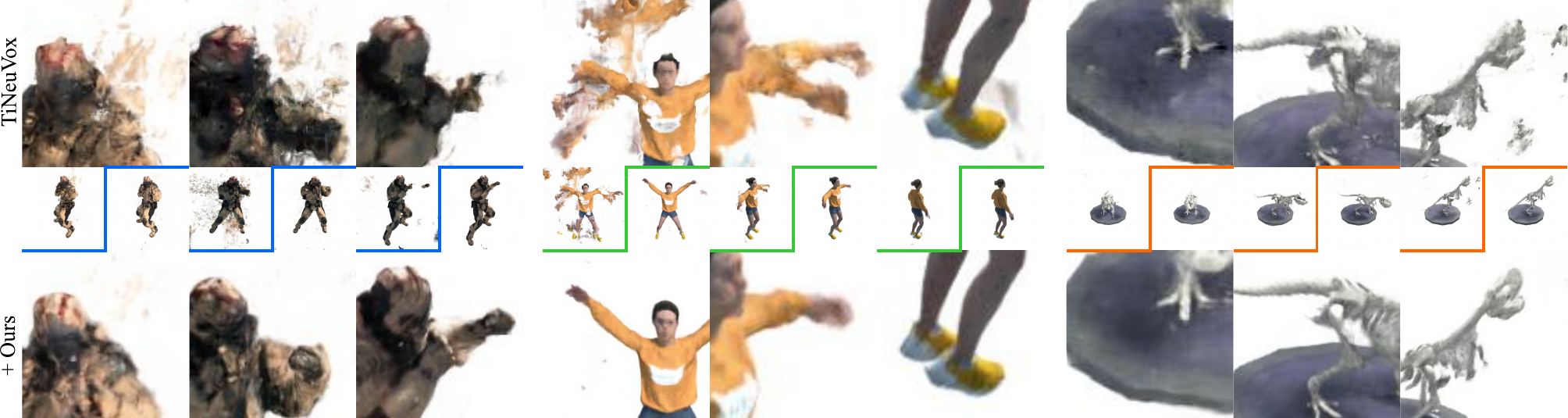}
\caption{Qualitative comparisons on dynamic scenes. TiNeuVox \cite{tineuvox} uses both low- and high-frequency by default, thus overfitting few-shot inputs. After adopting our method, the images have finer details and fewer floating outliers.}\label{fig:dynamic}
\vspace{-1em}
\end{figure*}

\subsection{Forward-facing scenes}
We evaluate our framework on real-world forward-facing images in this section. Four corner views are selected as the training view for each scene.
Note that the camera intrinsics and extrinsics are not provided in StanfordLF, so to test the performance of NeRF on the dataset StanfordLF, we use an EPI-based parameterization in NeRF. A more detailed definition of the coordinates is included in the supplementary.
Besides the two-plane light field dataset, we also compare our method with NeRF on the LLFF dataset \cite{mildenhall2019llff}.
In \cref{tab:real-images}, we compare our method with NeRF. 
\cref{fig:image-compare} visualizes the rendered images from our method and NeRF. The top two rows of images are from the StanfordLF dataset, and the third row is from the LLFF dataset. 
The images in this section are with small baselines. Therefore the vanilla NeRF can generate decent novel views despite the access to only 4 corner views. We can observe from \cref{tab:real-images} and \cref{fig:image-compare} that NeRF will not overfit the training views and no floating outliers are presented with these inputs, harnessing low-frequency can still be beneficial for rendering realistic novel views.

\begin{table}
\centering
\resizebox*{\columnwidth}{!}{
\begin{tabular}{l|c|ccc}
\toprule
\textbf{Method} & \textbf{Time} (GPU/h) $\downarrow$ & \textbf{PSNR}$\uparrow$ & \textbf{SSIM}$\uparrow$ & \textbf{LPIPS}$\downarrow$ \\\midrule
TiNeuVox-S \cite{tineuvox} & 0.17 & 21.392 & 0.864 & 0.182 \\
DietNeRF \cite{Jain_2021_ICCV} & 4.25 & 22.487 & 0.873 & 0.177 \\
DietNeRF-ft \cite{Jain_2021_ICCV} & 5.17 & 22.815 & 0.878 & 0.171 \\
RegNeRF \cite{niemeyer2021regnerf} & 3.33 & 23.995 & 0.889 & 0.143 \\
\cmidrule(lr){1-5}
HALO (\emph{Ours}) & 0.67 & \textbf{25.687} & \textbf{0.914} & \textbf{0.116} \\
\bottomrule
\end{tabular}
}
\caption{Quantitative comparisons on dynamic scenes. Our method can be easily extended to extra input dimension (\ie, time $t$), while other methods are applied frame-by-frame. Time indicates the training time on one scene. Our method does not require querying a pre-trained prior network, thus keeping efficient.}\label{tab:dynamic}
\end{table}

\begin{figure}
\includegraphics[width=\columnwidth,trim={0 0 0 2em},clip]{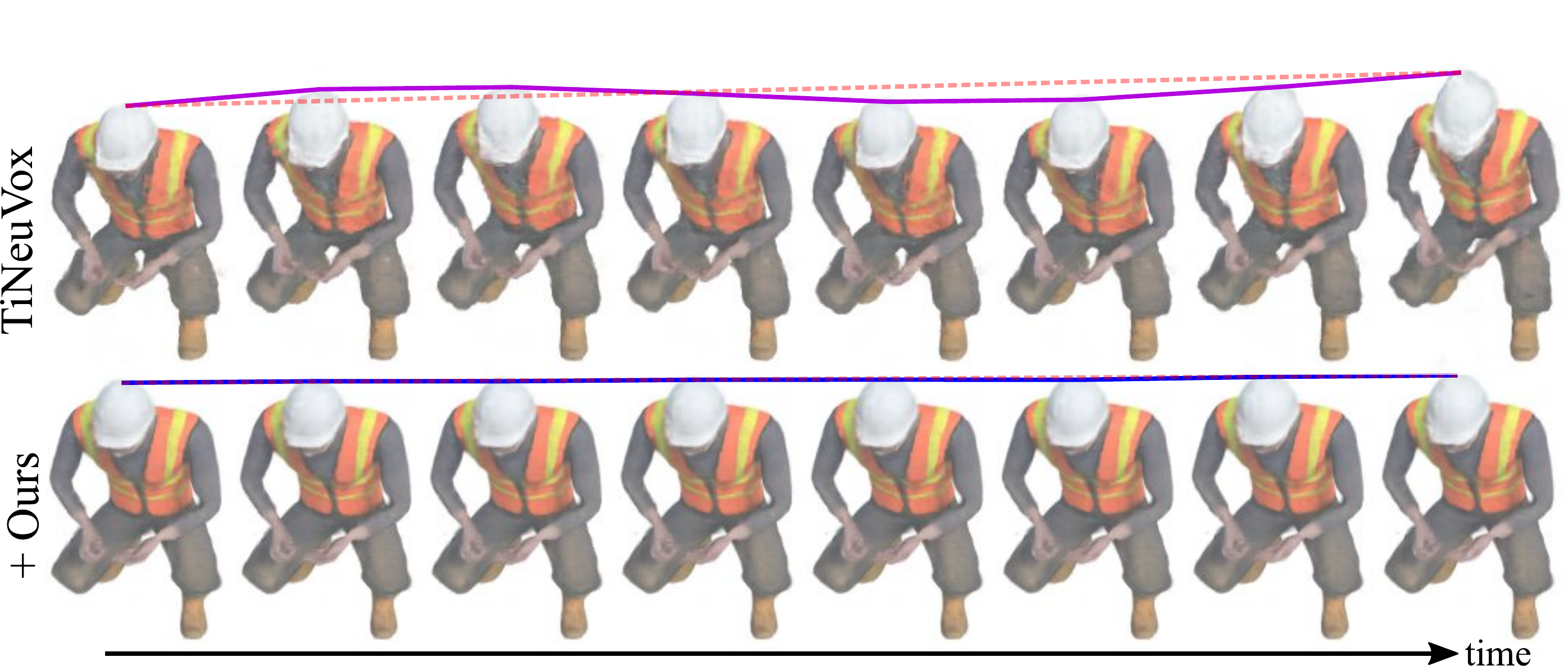}
\caption{Temporal flickering motion is observed on TiNeuVox under few-shot dynamic settings since the default frequency of time $t$ is high. Solid lines demonstrate the trajectory and dashed lines demonstrate the start-to-end linear trajectory.}\label{fig:dynamicmoving}
\vspace{-1em}
\end{figure}

\subsection{Dynamic scenes}
Our method uses low-frequency inputs as a regularization; thus, extra dimension in the input space can be readily implemented. We consider an extra time dimension and test our method on dynamic scenes. For dynamic scenes, not only are multiview observations sparse, sampling on the time axis becomes sparse as well for few-show challenges. We adopt the recent efficient reconstruction method TiNeuVox \cite{tineuvox} as the baseline, which can be trained within minutes. Like in vanilla NeRF, low-frequency reconstruction can be achieved by reducing the frequency in positional encoding. 
We skip training a low-frequency ray-based field in this experiment since directly querying from a low-frequency TiNeuVox is cheap. When tuning the low-frequency input, we use a similar strategy as on static scenes, but with a modification that renders new images with a fixed timestamp.

In \cref{tab:dynamic}, we present the quantitative results on the dynamic setting. TiNeuVox-S is adopted as the base model for all the experiments. TiNeuVox uses both low- and high-frequency for reconstruction by default, while low-frequency inputs regularize ours. We can observe that our method consistently improves performance when observations are sparse. We visualize rendered images in \cref{fig:dynamic}, from which we can observe clear visual improvements after adopting our regularization method.
Our method does not require data-driven priors to regularize, therefore keeping the efficiency of volume based NeRF models.
Furthermore, we observe that the high-frequency of the time axis introduces temporal flickering motion when observations of dynamic sequence are sparse. An illustration is demonstrated in \cref{fig:dynamicmoving}, and more can be found in our video.


\section{Conclusion}
We propose to harness the low-frequency NeRF and leverage it to regularize the high-frequency NeRF so that it will not overfit under the few-shot setting. The regularization is conducted in the input space so our method can be readily applied to static and dynamic scenes. 
Also benefiting from the different interpolation and extrapolation properties of low- and high-frequency, our method can extrapolate periodic contents and render realistic novel views on unobserved areas.
Furthermore, we design a simple-yet-effective criterion for determining a NeRF's frequency to avoid overfitting. Our experimental results demonstrate the effectiveness of our proposed solution's effectiveness and the potential of harnessing low-frequency neural fields. 


{\small
\bibliographystyle{ieee_fullname}
\bibliography{egbib}
}

\clearpage
\onecolumn
\appendix
{\centering\textbf{\huge Appendix}}

\section{Extend to light field data}
\begin{figure}
    \centering
    \includegraphics[width=0.5\columnwidth,trim={0 0 0 0},clip]{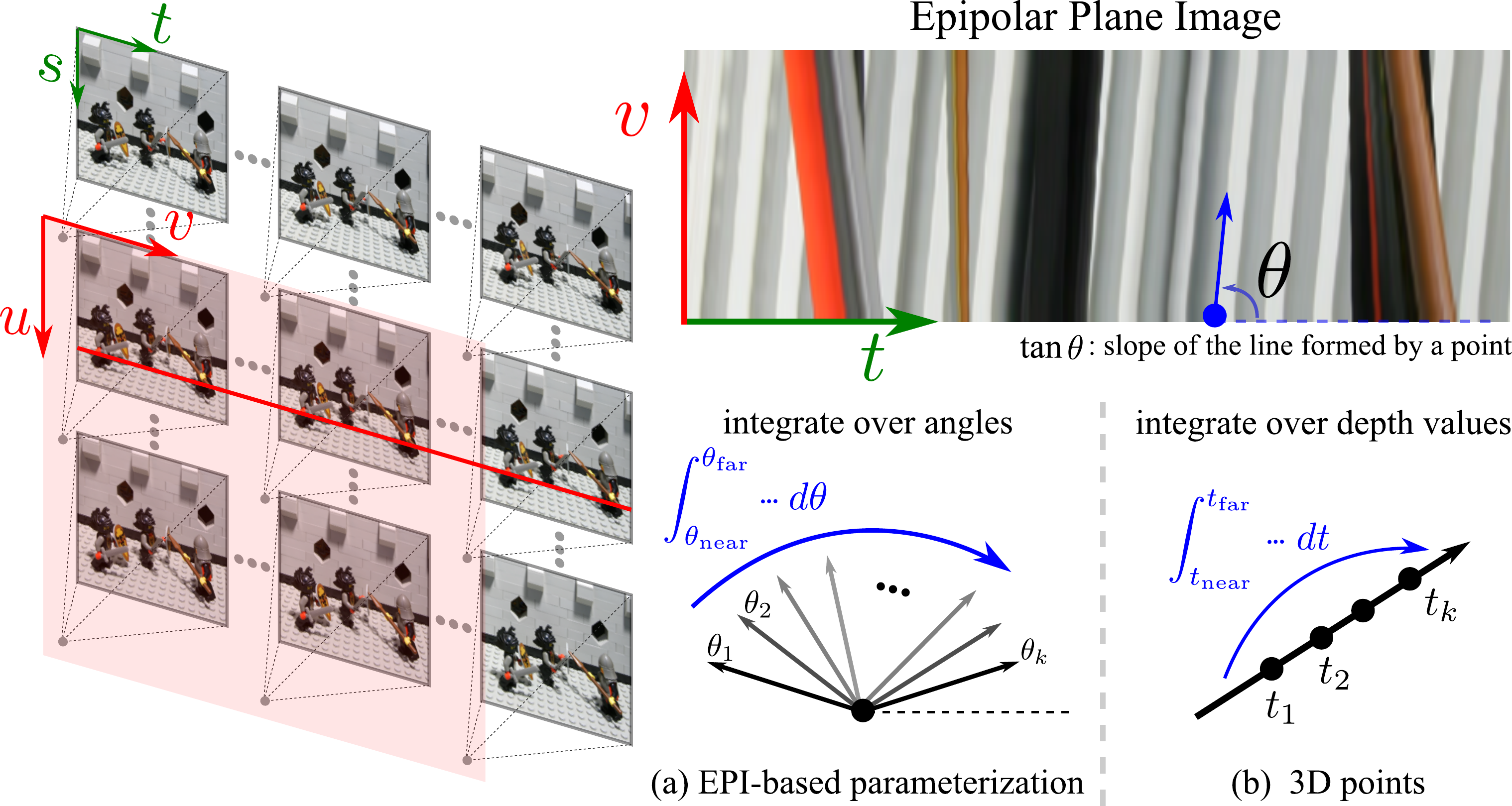}
    \caption{Illustration of Epipolar Plane Images (EPIs). Each point in the space forms a line on EPIs. To accumulate the color of a ray, integration is taken over (a) the angle $\theta$ for the EPI-based coordinates and (b) depth $t$ for point-based coordinates.}
    \label{fig:epi-illu}
\end{figure}

To further validate the proposed framework, we consider the angular-spatial resolution tradeoff on two-plane parameterized light field data. Though we can estimate camera intrinsics and extrinsics from images, directly processing light rays is a more general approach and sometimes favorable. We design a parameterization for the 3D points in the space, which enables controlling the geometry with light ray inputs.
\paragraph{EPI-based parameterization.}
The two-plane parameterization light field \cite{LevoyH96,GortlerGSC96} is a well-established method for representing rays in the space. 
Each ray is parameterized by the intersection points on the camera plane $uv$ and the image plane $st$.
Then, a point in space will be a line on the $us$ or $st$ slice of the light field (\cref{fig:epi-illu}), which is also known as Epipolar Plane Images (EPIs) \cite{bolles1987epipolar}.
We denote the slope of the line as $tan\theta$, then a 3D point can be determined by $(u,v,s,t,\theta)$. Consequently, points on the ray $(u,v,s,t)$ can be represented by $\{(u,v,s,t,\theta_i)\}_{i=0}^{i=N}$, where $N$ is the number of sampled points. To get the color of the ray, we take the integral over $\theta$. 

\begin{figure}
    \centering
    \includegraphics[width=0.5\columnwidth,trim={0 0 0 0},clip]{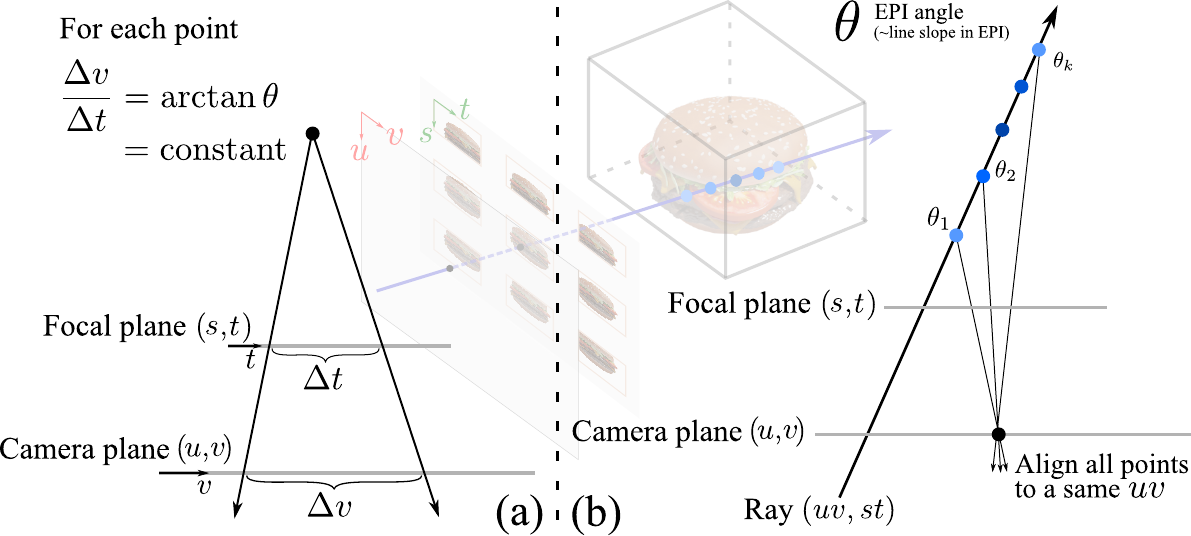}
    \caption{Points are parameterized by $(s,t,\theta)$. (a) Once the two planes are determined, $\frac{\Delta v}{\Delta t}$ is a constant. (b) We align all rays to the same $uv$ for representing points with 3D vectors.}
    \label{fig:epi-points}
\end{figure}

The parameterization $(u,v,s,t,\theta)$ for 3D points is redundant and can be simplified. In \cref{fig:epi-points}, we demonstrate how $v$ and $t$ are correlated: For the same 3D point, two rays observing the point will satisfy $ \frac{\Delta v}{\Delta t}=\arctan\theta=\mathrm{constant}$ as the two planes are fixed.
Based on this fact, we propose to align all $(u,v,s,t,\theta)$ to a fixed $uv$ for representing the spatial location of a 3D point. 
Let the $u^*$ and $v^*$ be the fixed value, then $s'=s+\frac{u-u^*}{\arctan\theta}$ and $t'=t+\frac{v-v^*}{\arctan\theta}$.
In this way, $(u,v,s,t,\theta)$ and $(u^*,v^*,s',t',\theta)$ represent the same 3D point. After aligning all 3D points to the same $u^*v^*$, each point is now parameterized by a 3D vector $(s',t',\theta)$. The inputs of radiance field are then switched to $(s',t',\theta)$ for 3D points and $(u,v)$ for viewing direction.

\paragraph{Joint training}
The two-stage training scheme limits the framework's extensibility since the low-frequency cannot be further optimized once finished. For example, a motion field which is widely used for modeling dynamic scenes (\eg, \cite{gafni2020dynamic,pumarola2020d,li2020neural,Tretschk_2021_ICCV,li2021neural,Park_2021_ICCV}) cannot be directly adopted into the two-stage framework.

\begin{figure}
    \centering
    \includegraphics[width=0.5\columnwidth,trim={0 0 0 0},clip]{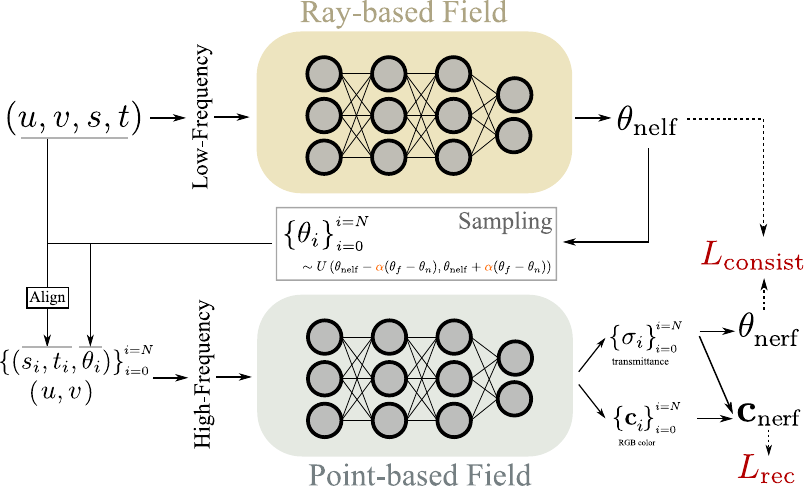}
    \caption{The overall framework of harnessing low-frequency neural fields. Inputs from 4D light field data are used for demonstration.}
    \label{fig:epi-train}
\end{figure}

An overview of our proposed framework for harnessing low-frequency neural fields is illustrated in \cref{fig:epi-train}. 
Inputs from 4D light field data are used for demonstration. For rendering with general 3D world coordinates (\ie, with known intrinsics and extrinsics), there are two small differences: First, the inputs for points are now $(x,y,z)$ and the align procedure is not needed; Second, the target for $L_{\mathrm{consist}}$ is from another Lo-NeRF since the baselines are much larger than light field data.

For each ray $(u,v,s,t)$, the ray-based field outputs a $\theta_{\mathrm{ray}}$, which is then used for guiding the sampling of $N$ points with $\{\theta_i\}_{i=0}^{i=N}$ on the ray. Each $\theta_i$ is uniformly sampled within a range dynamically adjusted along with training. 
On light field data, the ray-based field and the point-based field are jointly optimized. A consistent loss is adopted to update the ray-based field. The difference is that the ray-based field can be directly updated along with Hi-NeRF since the training is more stable with sparse observations on forward-facing scenes. Joint training the two fields is implemented by progressively regularizing the high-frequency radiance field, making the framework more compact and efficient. 
Mathematically, we set $\theta_i\sim U\left(\theta_{\mathrm{ray}}-\alpha(\theta_f-\theta_n),\theta_{\mathrm{ray}}+\alpha(\theta_f-\theta_n)\right)$, where $\alpha\in[k,1]$ and $k$ is the defined range for sampling. During training, $\alpha$ is 1 at the beginning and then linearly converge to $k$. In this way, the geometry of the neural radiance field will gradually converge to local details.

\begin{figure}
    \centering
    \includegraphics[width=0.5\columnwidth,trim={0 0 0 0},clip]{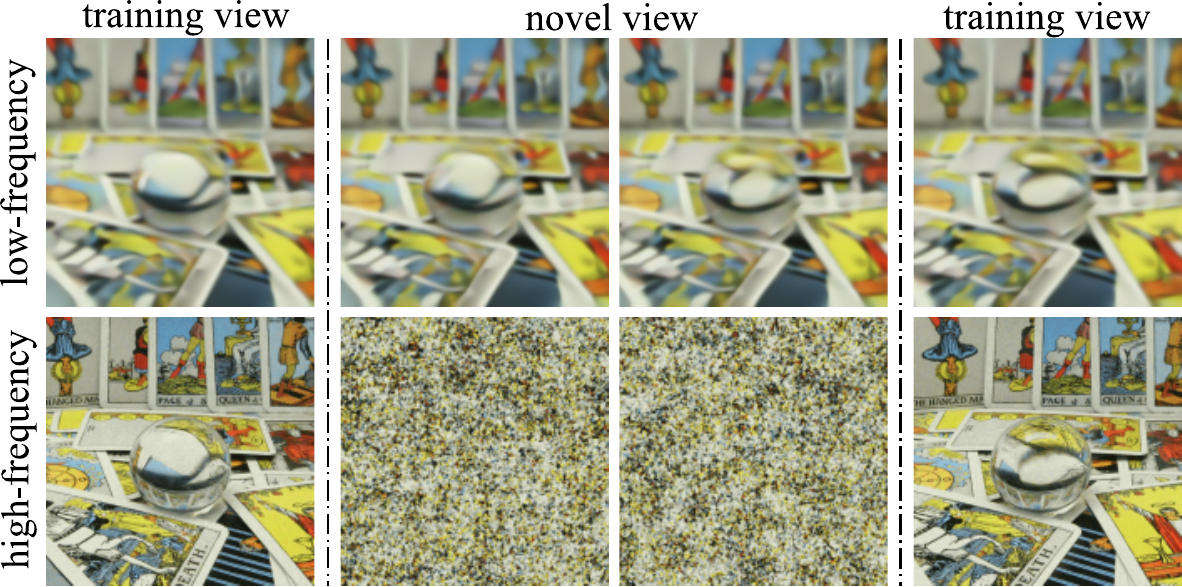}
    \caption{Rendering results of neural light fields with low- and high-frequency inputs on two adjacent training views and novel interpolation views. Using low-frequency leads to smooth color change across rays while high-frequency leads to overfitting.}
    \label{fig:lf-nelf}
    \vspace{-8pt}
\end{figure}

Besides the reconstruction loss for supervising the radiance field, a consistent loss is adopted to update the neural light field. The consistent loss is defined as the difference between $\theta$ from the neural light field and the neural radiance field, that is, $L_{\mathrm{consist}}=\|\theta_{\mathrm{nelf}}-\theta_{\mathrm{nerf}}\|_2^2$. Note that the neural light field takes low-frequency inputs, aiming for regularizing the geometry of the high-frequency radiance field. A neural light field with low-frequency inputs leads to similar effects as a neural radiance field: lacking details but generalizing well. 

\section{Implementation details}
Our method is implemented with PyTorch \cite{paszke2019pytorch}. Each model is trained on one 2080Ti with batch size 3072. Important hyperparameters are listed below for each experiment and other detailed training hyperparameters can be find in our code.
\paragraph{2D toy demonstration.}
The low-frequency neural field is with $L=5$ and $s=5$. The high-frequency neural field is with $L=10$ and $s=5$.
\paragraph{Experiments with 8 views.}
Training images for the 8 view setting are included in \cref{tab:8view}. For the low-frequency P.E., we set $L=5$ and an extra scale parameter $s=32$. The scale parameter indicates that all inputs are divided by $s$. The low-frequency radiance field is trained for 8,000 iterations and the high-frequency radiance field is trained for 120,000 iterations.
\paragraph{Experiments with 14 one side views.}
Training images are 
\begin{quote}
\texttt{[`r\_58.png', `r\_5.png', `r\_2.png', `r\_8.png', `r\_9.png', \\`r\_10.png', `r\_16.png', `r\_34.png', `r\_35.png', `r\_40.png', \\`r\_52.png', `r\_53.png', `r\_54.png', `r\_60.png']}. 
\end{quote}
We set $s=1$ and $L=5$ for the low-frequency neural fields.
\paragraph{Experiments on light field data.}
For light field data, we use the Normal distribution based P.E. proposed by \cite{TancikSMFRSRBN20}. 
The std for $uv$ and $st$ are 64 with length 10, so $uv$ is the low-frequency part. The std for $\theta$ is 8 with length 5.
The parameter $\alpha$ linearly decay to 0.5 after 20,000 iterations. 
\paragraph{Experiments on LLFF \cite{mildenhall2019llff}.}
The four training views and one validation view on the four selected scenes are as in \cref{tab:llff}.
\begin{table*}[t]
\centering
\begin{tabular}{ccc}
\toprule
Scene & Train & Val\\ \midrule
flower & \texttt{['007.png', '008.png', '011.png', '009.png']} & \texttt{['010.png']} \\
leaves & \texttt{['004.png', '005.png', '008.png', '006.png']} & \texttt{['007.png']} \\
orchids & \texttt{['008.png', '006.png', '014.png', '015.png']} & \texttt{['007.png']} \\
fern & \texttt{['005.png', '013.png', '015.png', '016.png']} & \texttt{['014.png']} \\
\bottomrule
\end{tabular}
\caption{Four training views and one validation view on LLFF.}
\label{tab:llff}
\end{table*}

\section{Additional results}
First, detailed results on each scene are included in \cref{tab:8view-detail} (for Tab 1 in the main text), \cref{tab:lightfield-detail} (for Tab 3 in the main text) and \cref{tab:llff-detail} (for Tab 3 in the main text).

\begin{figure*}[t]
\centering
\includegraphics[width=0.8\textwidth]{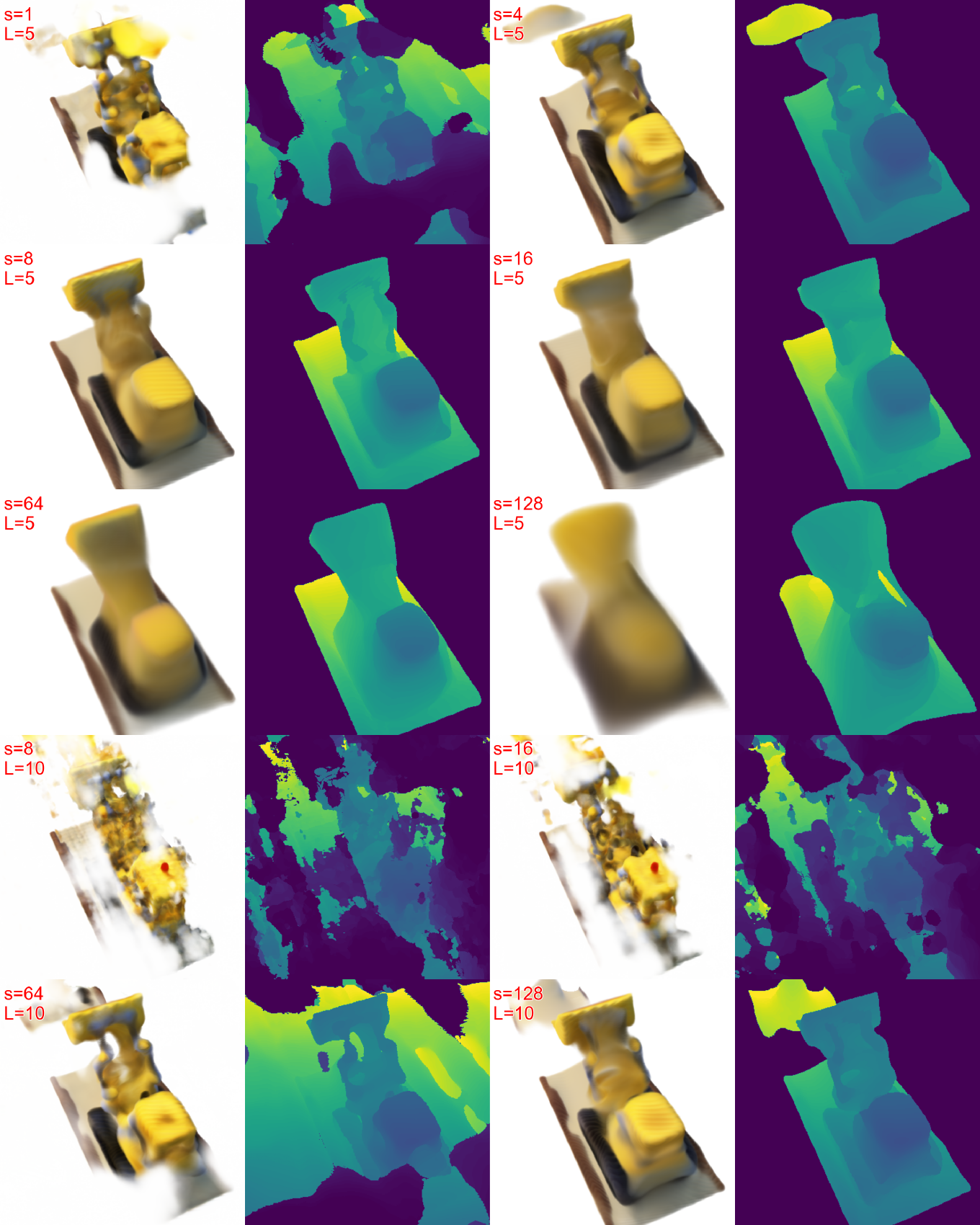}
\caption{Results with different frequency settings (higher $s$ and $L$ indicates a higher frequency).}
\label{fig:freq}
\end{figure*}

\begin{figure*}
\centering
\includegraphics[width=0.8\textwidth]{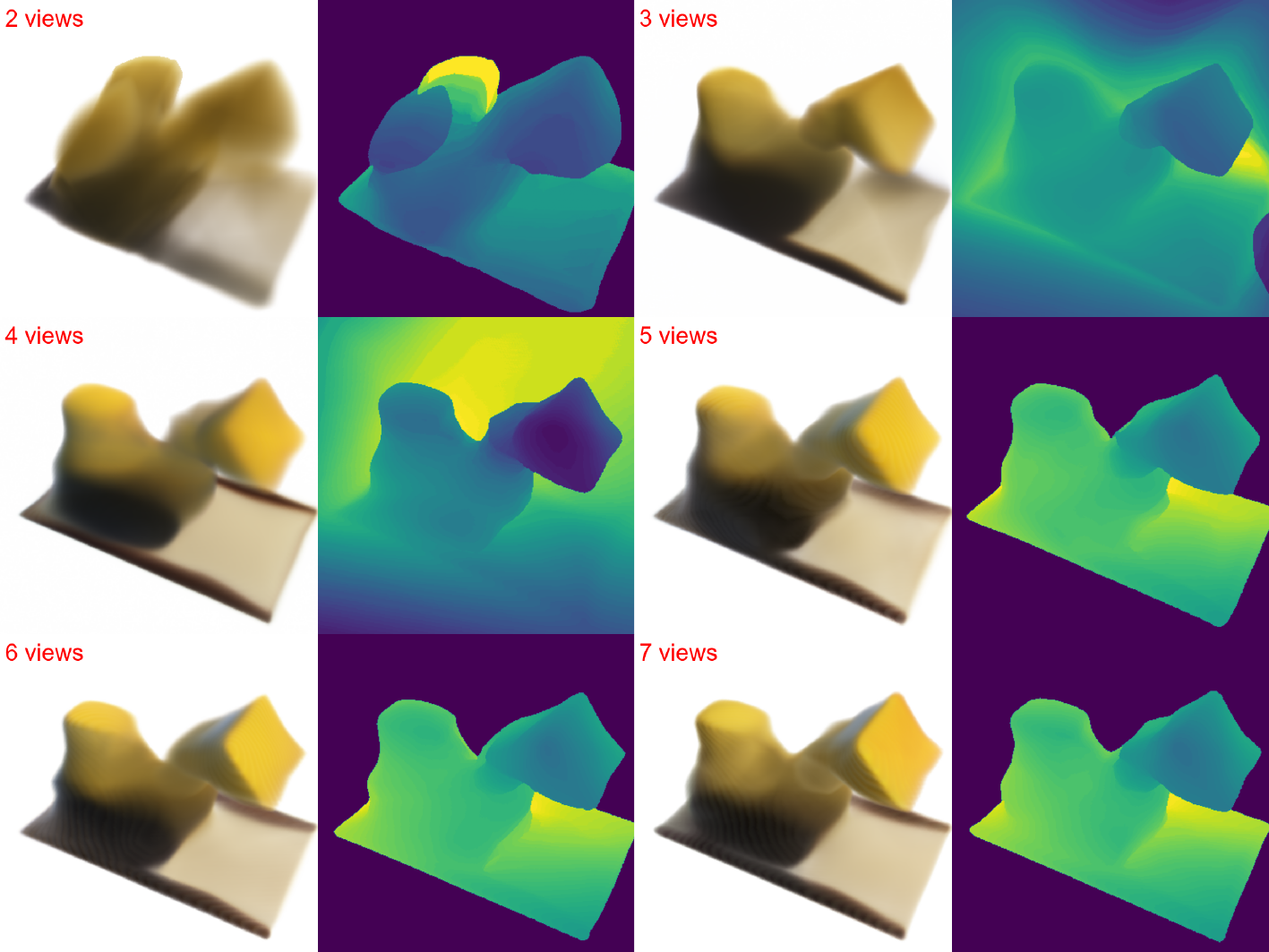}
\caption{Results with different number of views.}
\label{fig:views}
\end{figure*}

Two ablation studies are conducted. \cref{fig:freq} demonstrates the impact of different frequency settings. As we admitted in the limitation section, the best frequency for a smooth geometry needs to be determined empirically. Fortunately, as demonstrated by \cref{fig:freq}, it not hard to find an appropriate frequency for smooth effects. Also, in \cref{fig:views}, we show the impact of the number of input views. First $k$ views demonstrated in \cref{tab:8view-detail} are used for the $k$ view setting. It can be observed that two views are not able to generate a good rough geometry, which is another limitation pointed out in the main text. Still, it is interesting to observe that 3 views are enough for a good rough geometry, though some areas are filled with undesired white points.

Per-scene performance on the dynamic scenes (D-NeRF dataset \cite{pumarola2020d}) is presented in \cref{tab:dynamic-per-scene}.

\begin{table*}
\centering
\resizebox{\textwidth}{!}{
\begin{tabular}{cc}
\toprule
Scene & Images \\
\midrule
lego & \texttt{['r\_2.png', 'r\_16.png', 'r\_93.png', 'r\_55.png', 'r\_73.png', 'r\_86.png', 'r\_26.png', 'r\_75.png']} \\
chair & \texttt{['r\_86.png', 'r\_73.png', 'r\_26.png', 'r\_2.png', 'r\_55.png', 'r\_93.png', 'r\_16.png', 'r\_75.png']} \\
drums & \texttt{['r\_86.png', 'r\_93.png', 'r\_75.png', 'r\_26.png', 'r\_55.png', 'r\_73.png', 'r\_16.png', 'r\_2.png']} \\
ficus & \texttt{['r\_2.png', 'r\_93.png', 'r\_73.png', 'r\_86.png', 'r\_75.png', 'r\_26.png', 'r\_55.png', 'r\_16.png']} \\
mic & \texttt{['r\_55.png', 'r\_2.png', 'r\_93.png', 'r\_75.png', 'r\_16.png', 'r\_26.png', 'r\_86.png', 'r\_73.png']} \\
ship & \texttt{['r\_55.png', 'r\_93.png', 'r\_26.png', 'r\_75.png', 'r\_16.png', 'r\_33.png', 'r\_73.png', 'r\_86.png']} \\
materials & \texttt{['r\_75.png', 'r\_26.png', 'r\_93.png', 'r\_55.png', 'r\_86.png', 'r\_16.png', 'r\_2.png', 'r\_73.png']} \\
hotdog & \texttt{['r\_16.png', 'r\_93.png', 'r\_75.png', 'r\_86.png', 'r\_2.png', 'r\_55.png', 'r\_73.png', 'r\_26.png']} \\
\bottomrule
\end{tabular}
}
\caption{Training images for the 8 view setting.}
\label{tab:8view}
\end{table*}


\begin{table*}
\centering
\begin{tabular}{lcccccccc}
\toprule
PSNR$\uparrow$ & Lego & Chair & Drums & Ficus & Mic & Ship & Materials & Hotdog \\
\midrule
NeRF & 9.726 & 21.049 & 17.472 & 13.728 & 26.287 & 12.929 & 7.837 & 10.446 \\
NV & 17.652 & 20.515 & 16.271 & 19.448 & 18.323 & 14.457 & 16.846 & 19.361 \\
Simplified NeRF & 16.735 & 21.870 & 15.021 & 21.091 & 24.206 & 17.092 & 20.659 & 24.060 \\
DietNeRF & 23.897 & 24.633 & 20.034 & 20.744 & 26.321 & 23.043 & 21.254 & 25.250 \\
DietNeRF ft & 24.311 & 25.595 & 20.029 & 20.940 & 26.794 & 22.536 & 21.621 & 26.626 \\
HALO (\emph{Ours})&
23.927 & 25.203 & 19.543 & 21.531 & 25.650 & 22.223 & 21.163 & 26.909 \\
\cmidrule(lr){1-9}
\textcolor{gray}{NeRF, 100 views} & \textcolor{gray}{31.618} & \textcolor{gray}{34.073} & \textcolor{gray}{25.530} & \textcolor{gray}{29.163} & \textcolor{gray}{33.197} & \textcolor{gray}{29.407} &\textcolor{gray}{29.340} & \textcolor{gray}{36.899} 
\\
\bottomrule
\vspace{6pt}\\
\toprule
SSIM$\uparrow$ & Lego & Chair & Drums & Ficus & Mic & Ship & Materials & Hotdog \\
\midrule
NeRF & 0.526 & 0.861 & 0.770 & 0.661 & 0.944 & 0.605 & 0.484 & 0.644 \\
NV & 0.707 & 0.795 & 0.675 & 0.815 & 0.816 & 0.602 & 0.721 & 0.796 \\
Simplified NeRF & 0.775 & 0.859 & 0.727 & 0.872 & 0.930 & 0.694 & 0.823 & 0.894 \\
DietNeRF & 0.863 & 0.898 & 0.843 & 0.872 & 0.944 & 0.758 & 0.843 & 0.904 \\
DietNeRF ft & 0.875 & 0.912 & 0.845 & 0.874 & 0.950 & 0.757 & 0.851 & 0.924 \\
HALO (\emph{Ours})&
0.854 & 0.898 & 0.818 & 0.888 & 0.936 & 0.756 & 0.826 & 0.930 \\
\cmidrule(lr){1-9}
\textcolor{gray}{NeRF, 100 views} & \textcolor{gray}{0.965} & \textcolor{gray}{0.978} & \textcolor{gray}{0.929} & \textcolor{gray}{0.966} & \textcolor{gray}{0.979} & \textcolor{gray}{0.875} & \textcolor{gray}{0.958} & \textcolor{gray}{0.981} \\
\bottomrule
\vspace{6pt}\\
\toprule
LPIPS$\downarrow$ & Lego & Chair & Drums & Ficus & Mic & Ship & Materials & Hotdog \\
\midrule
NeRF & 0.467 & 0.163 & 0.231 & 0.354 & 0.067 & 0.375 & 0.467 & 0.422 \\
NV & 0.253 & 0.175 & 0.299 & 0.156 & 0.193 & 0.456 & 0.223 & 0.203 \\
Simplified NeRF & 0.218 & 0.152 & 0.280 & 0.132 & 0.080 & 0.283 & 0.151 & 0.139 \\
DietNeRF & 0.110 & 0.092 & 0.117 & 0.097 & 0.053 & 0.204 & 0.102 & 0.097 \\
DietNeRF ft & 0.096 & 0.077 & 0.117 & 0.094 & 0.043 & 0.193 & 0.095 & 0.067 \\
HALO (\emph{Ours})&
0.140 & 0.108 & 0.190 & 0.140 & 0.088 & 0.259 & 0.202 & 0.091 \\
\cmidrule(lr){1-9}
\textcolor{gray}{NeRF, 100 views} & \textcolor{gray}{0.033} & \textcolor{gray}{0.025} & \textcolor{gray}{0.064} & \textcolor{gray}{0.035} & \textcolor{gray}{0.023} & \textcolor{gray}{0.125} & \textcolor{gray}{0.037} & \textcolor{gray}{0.025} \\
\bottomrule
\end{tabular}
\caption{Detailed results on each scene for the 8 views setting.}
\label{tab:8view-detail}
\end{table*}

\begin{table*}
\centering
\begin{tabular}{lcccccc}
\toprule
& \multicolumn{3}{c}{NeRF} & \multicolumn{3}{c}{HALO} \\\cmidrule(lr){2-4}\cmidrule(lr){5-7}
Scene
& PSNR$\uparrow$ & SSIM$\uparrow$ & LPIPS$\downarrow$ 
& PSNR$\uparrow$ & SSIM$\uparrow$ & LPIPS$\downarrow$ \\ 
\midrule
amethyst & 23.146 & 0.708 & 0.342 & 27.088 & 0.797 & 0.262 \\
bracelet & 22.215 & 0.794 & 0.299 & 28.929 & 0.936 & 0.188 \\
cards-big & 17.485 & 0.535 & 0.507 & 20.613 & 0.778 & 0.260 \\
cards-small & 23.199 & 0.868 & 0.163 & 28.382 & 0.938 & 0.102 \\
chess & 27.707 & 0.858 & 0.337 & 33.682 & 0.946 & 0.227 \\
eucalyptus-flowers & 35.491 & 0.930 & 0.277 & 38.163 & 0.950 & 0.233 \\
jellybeans & 35.297 & 0.974 & 0.169 & 39.584 & 0.982 & 0.153 \\
lego-bulldozer & 24.700 & 0.774 & 0.482 & 28.688 & 0.894 & 0.330 \\
lego-gantry & 17.809 & 0.625 & 0.433 & 19.741 & 0.672 & 0.523 \\
lego-knights & 27.210 & 0.899 & 0.170 & 32.494 & 0.950 & 0.112 \\
lego-truck & 35.099 & 0.945 & 0.256 & 38.053 & 0.960 & 0.224 \\
stanfordbunny & 39.046 & 0.945 & 0.215 & 42.146 & 0.967 & 0.198 \\
treasure & 14.671 & 0.218 & 0.511 & 29.127 & 0.895 & 0.242 \\ 
\cmidrule(lr){1-7}
mean & 26.390 & 0.775 & 0.320 & 31.283 & 0.897 & 0.234 \\
\bottomrule
\end{tabular}
\caption{Detailed results on each scene for the light field data StanfordLF.}
\label{tab:lightfield-detail}
\end{table*}


\begin{table*}[t]
\centering
\resizebox{\textwidth}{!}{
\begin{tabular}{lccccccccccccccc}
\toprule
& \multicolumn{5}{c}{PSNR$\uparrow$} & \multicolumn{5}{c}{SSIM$\uparrow$} & \multicolumn{5}{c}{LPIPS$\downarrow$} \\
\cmidrule(lr){2-6}\cmidrule(lr){7-11}\cmidrule(lr){12-16}
& NeRF & IBRNet & MVSNeRF & \begin{tabular}{c}HALO\\w/o j.t\end{tabular} & HALO 
& NeRF & IBRNet & MVSNeRF & \begin{tabular}{c}HALO\\w/o j.t\end{tabular} & HALO
& NeRF & IBRNet & MVSNeRF & \begin{tabular}{c}HALO\\w/o j.t\end{tabular} & HALO \\ 
\midrule
fern & 23.995 & 23.225 & 23.723 & 22.64 & 23.10
& 0.763 & 0.716 & 0.727 & 0.774 & 0.795
& 0.274 & 0.302 & 0.299 & 0.266 & 0.253 \\
flower & 22.247 & 23.052 & 23.372 & 26.55  & 27.23
& 0.739 & 0.698 & 0.719 & 0.909 & 0.912
& 0.282 & 0.251 & 0.233 & 0.146 & 0.143 \\
leaves & 18.542 & 18.365 & 18.694 & 22.07  & 21.54
& 0.652 & 0.561 & 0.603 & 0.843 & 0.826
& 0.296 & 0.371 & 0.331 & 0.180 & 0.222 \\
orchids & 14.844 & 16.133 & 16.522 & 19.01 & 20.51
& 0.383 & 0.343 & 0.458 & 0.705 & 0.732 
& 0.508 & 0.559 & 0.425 & 0.286 & 0.258 \\
\cmidrule(lr){1-16}
mean & 19.907 & 20.194 & 20.578 & 22.567 & 23.095
& 0.634 & 0.580 & 0.627 & 0.808 & 0.816
& 0.340 & 0.371 & 0.322 & 0.219 & 0.219 \\
\bottomrule
\end{tabular}
}
\caption{Detailed results on each scene for the 4 views setting on LLFF.}
\label{tab:llff-detail}
\end{table*}

\begin{table*}[t]
\centering
\begin{tabular}{c|cc|cc|cc}
    \toprule
    \multirow{2}{*}{Scene} & \multicolumn{2}{c|}{PSNR$\uparrow$} & \multicolumn{2}{c|}{SSIM$\uparrow$} & \multicolumn{2}{c}{LPIPS$\downarrow$}\\
    & {\footnotesize TNV\cite{tineuvox}} & {\footnotesize + Ours} & {\footnotesize TNV\cite{tineuvox}} & {\footnotesize + Ours} & {\footnotesize TNV\cite{tineuvox}} & {\footnotesize + Ours} \\
    \midrule
    \textsc{\footnotesize Hellwarrior} & 12.288 & \textbf{17.558} & 0.733 & \textbf{0.832} & 0.349 & \textbf{0.247} \\
    \textsc{\footnotesize Bouncingballs} & 25.431 & \textbf{24.168} & 0.942 & \textbf{0.911} & 0.135 & \textbf{0.116} \\
    \textsc{\footnotesize Jumpingjacks} & 22.591 & \textbf{31.952} & 0.915 & \textbf{0.972} & 0.114 & \textbf{0.087} \\
    \textsc{\footnotesize Hook} & 16.676 & \textbf{27.065} & 0.692 & \textbf{0.953} & 0.370 & \textbf{0.064} \\
    \textsc{\footnotesize Lego} & 19.824 & \textbf{20.083} & 0.845 & \textbf{0.787} & 0.203 & \textbf{0.244} \\
    \textsc{\footnotesize Mutant} & 26.154 & \textbf{28.953} & 0.935 & \textbf{0.952} & 0.077 & \textbf{0.054} \\
    \textsc{\footnotesize Standup} & 24.404 & \textbf{28.980} & 0.934 & \textbf{0.963} & 0.089 & \textbf{0.042} \\
    \textsc{\footnotesize Trex} & 23.770 & \textbf{26.736} & 0.912 & \textbf{0.939} & 0.120 & \textbf{0.076} \\
    \midrule
    \rowcolor[gray]{0.95}
    Average & 21.392 & \textbf{25.687} & 0.864 & \textbf{0.914} & 0.182 & \textbf{0.116} \\
    \bottomrule
\end{tabular}
\caption{Few-shot rendering on dynamic scenes. Every 8 training images (\ie, sparse spatiotemporal inputs) from the D-NeRF \cite{pumarola2020d} dataset are used.}\label{tab:dynamic-per-scene}
\end{table*}

\end{document}